\documentclass{article}

\usepackage{arxiv}

\usepackage[utf8]{inputenc} 
\usepackage[T1]{fontenc} 
\usepackage{hyperref} 
\usepackage{url} 
\usepackage{booktabs} 
\usepackage{amsfonts} 
\usepackage{nicefrac} 
\usepackage{microtype} 
\usepackage{lipsum}
\usepackage{graphicx}
\graphicspath{ {./images/} }

\usepackage{todonotes} 

\usepackage{tabularx,ragged2e}
\usepackage{appendix}
\usepackage{booktabs} 
\usepackage{graphicx} 
\usepackage{array}
\usepackage{subcaption}
\usepackage[affil-it]{authblk} 
\usepackage{refcount} 
\usepackage{lineno}

\title{Does Conceptual Representation Require Embodiment? Insights From Large Language Models}

\author[1,*]{Qihui Xu}
\author[2]{Yingying Peng}
\author[3]{Samuel A. Nastase}
\author[4,5]{Martin Chodorow}
\author[2]{Minghua Wu}
\author[2,*]{Ping Li}

\affil[1]{Basque Center on Cognition, Brain and Language}
\affil[2]{The Hong Kong Polytechnic University}
\affil[3]{Princeton Neuroscience Institute, Princeton University}
\affil[4]{Hunter College, City University of New York}
\affil[5]{Graduate Center, City University of New York}

\begin{document}
\maketitle
\begingroup\def\thefootnote{*}\footnotetext{Correspondence: Qihui Xu, q.xu@bcbl.eu; Ping Li, pi2li@polyu.edu.hk}\endgroup

\begin{abstract}
To what extent can language alone give rise to complex concepts, or is embodied experience essential? Recent advancements in large language models (LLMs) offer fresh perspectives on this question. Although LLMs are trained on restricted modalities, they exhibit human-like performance in diverse psychological tasks. Our study compared representations of 4,442 lexical concepts between humans and ChatGPTs (GPT-3.5 and GPT-4) across multiple dimensions, including five key domains: emotion, salience, mental visualization, sensory, and motor experience. We identify two main findings: 1) Both models strongly align with human representations in non-sensorimotor domains but lag in sensory and motor areas, with GPT-4 outperforming GPT-3.5; 2) GPT-4's gains are associated with its additional visual learning, which also appears to benefit related dimensions like haptics and imageability. These results highlight the limitations of language in isolation, and that the integration of diverse modalities of inputs leads to a more human-like conceptual representation.

\end{abstract}


\section{Introduction}
Imagine learning about the color \textit{red} without ever seeing a rose, a stop sign, or a sunset. Can we truly represent the concept \textit{red} in all its richness? This question invokes a longstanding debate about the interplay between physical experience and conceptual understanding. On one hand, theories of embodied cognition posit that our senses are our gateways to knowledge \cite{barsalou1999, barsalou2008grounded, glenberg2000symbol}; seeing `red' is integral to understanding it. On the other hand, opposing theories argue that the mind can form conceptually rich representations that are derived from language alone, independent of direct sensory experience. For example, studies show that individuals born blind can understand and respond to the concept of color similarly to those who can see \cite{wang2020, kim2021, bottini2020brain}. So, when sensory input is absent, to what extent can language alone inform our conceptual representation of the world? How indispensable is bodily experience in shaping our conceptual world? 

Disentangling the various sources for conceptual formation is challenging. While studies involving both congenitally blind and sighted individuals have provided valuable insights, they face several challenges: First, they often overlook the multi-dimensional nature of human conceptual representation \cite{banks2023multi, pexman2023social}. For instance, processing the concept of \textit{red} may evoke not just perceptions of the color, but also associated objects, emotions, inner feelings, etc. Second, the small scale of tested words constrains external validity, failing to account for the variety of concepts in daily use \cite{lenci2013blind, bottini2020brain, brysbaert2014concreteness}. Moreover, there can be potential knowledge transfer across domains. Even without visual input, individuals can tap into other sensory channels like touch and internal sensations, which have been shown to correlate with visual knowledge \cite{lynott2020lancaster}. 

Recent advances in large language models (LLMs) may offer a unique avenue to explore human conceptual representations. LLMs trained on massive amounts of data from limited modalities (e.g., text for GPT-3.5, and text and images for GPT-4) can be viewed as disembodied learners—analogous to subjects who do not receive multimodal inputs (in the case of GPT-3) and have no physical body to interact with the world (in the case of both GPT-3 and GPT-4). Remarkably, they nonetheless exhibit human-like performance in various cognitive tasks \cite{binz2023using, kosinski2023theory, cai2023does, marjieh2023language}. In the same way that large language models demonstrate the feasibility of learning syntactic structure from surface-level language exposure alone \cite{piantadosi2023modern}, they may also be used to evaluate the feasibility of learning physical, embodied features of the world from language alone \cite{phildeeplearning2023}. For example, some argues that multimodal experiences are necessary to enable humans to grasp concepts more efficiently, with far less linguistic exposure\cite{lake2023word, xiang2023language, chemero2023llms}, given that LLMs rely on immense volumes of text—for example, equivalent to 20,000 years of human reading for GPT-3\cite{warstadt2022artificial}. Nonetheless,  others argue that language itself can act as a surrogate `body' for these models (reminiscent of the largely conceptualized and disembodied color knowledge for the blind) \cite{patel2022mapping, marjieh2023language}. Therefore, the significance of using LLMs is highlighted by their ability to a) facilitate the examination of how different modalities (e.g., text language, image, audio, etc.) of input influence learning processes, thereby shedding light on the intricacies of language and cognition, and b) offer avenues for research that transcend the constraints inherent in human-based studies. These aspects have also been extensively discussed in recent opinion articles \cite{frank2023large, blank2023large}. However, the application of LLMs as cognitive models faces critics about the absence of stringent validation processes comparable to those in human subject research \cite{shapira2023clever, ma2023tomchallenges, frank2023large}. In contrast to human studies, evaluations of LLMs often lack comprehensive verification and can be sensitive to the phrasing of prompts \cite{ma2023tomchallenges}, which might lead to seemingly accurate yet superficial responses. Consequently, ensuring validity in the use of LLMs is imperative. When used with these considerations, LLMs hold immense potential as a means to enhance our comprehension of human cognition.

\begin{figure}[h]\centering    
    \includegraphics[width=\textwidth]{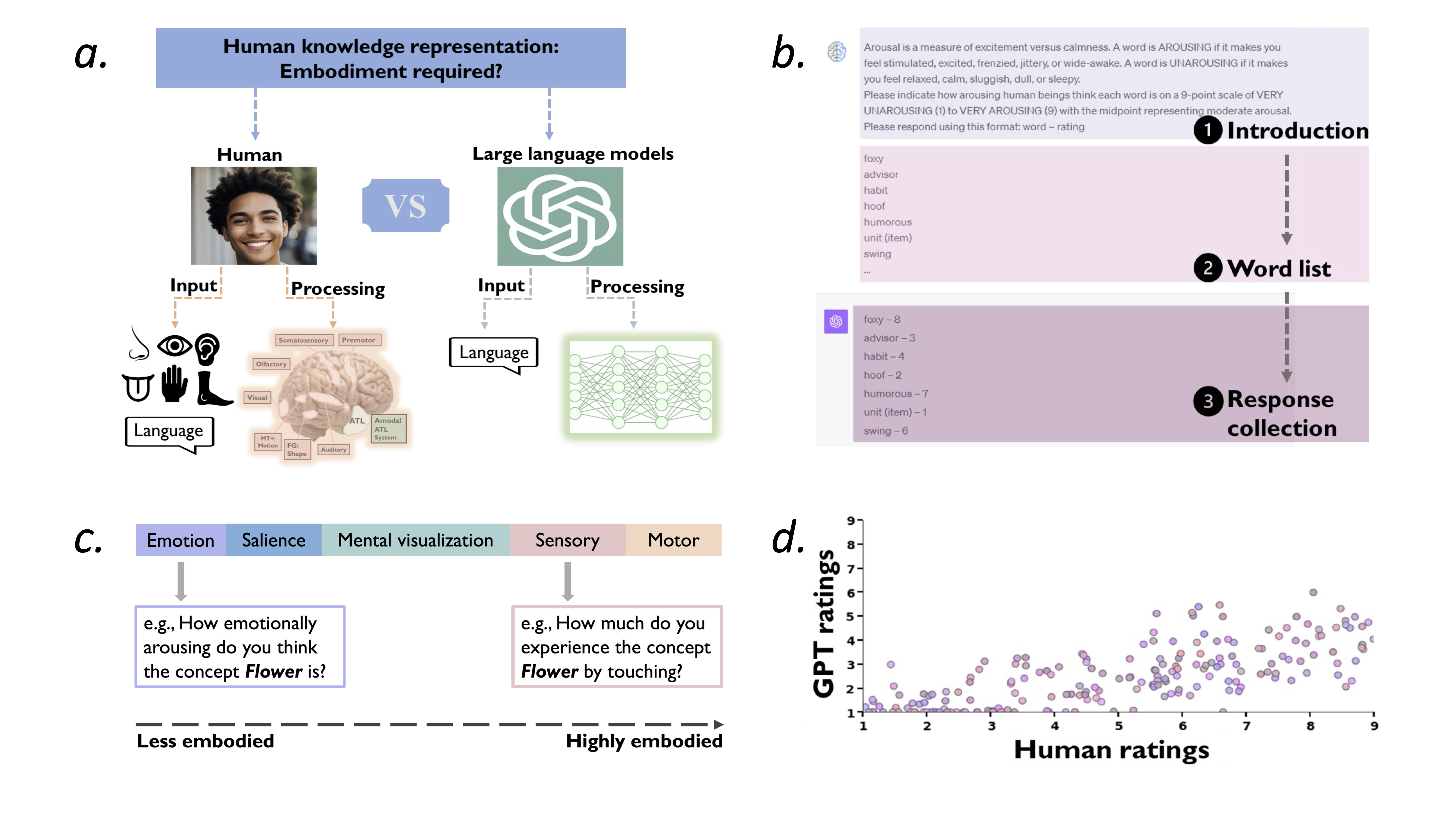}
    \caption{\textbf{a}. Schematic depiction of the research question and approach. This study aims to investigate the extent to which human conceptual representation requires grounding through embodied experience. \textbf{b}. The major categories of psycholinguistic dimensions studied, from less sensorimotor-involved dimensions (e.gs., emotional arousal) to more directly sensorimotor-involved dimensions (e.g., touching). \textbf{c}. Schematic of the ChatGPT testing procedure. The model prompt and design were aligned with the instructions for human subjects, which started with explaining the dimension and listing the words to be rated. ChatGPT would then provide ratings per word as required. \textbf{d}. Schematic of measuring the model's overall similarity with humans through correlation.}
    \label{overview}
\end{figure}

In this study, we probe the roles of language and embodied experience in shaping human conceptual knowledge, utilizing the potential of LLMs on modeling human cognition. We compared word ratings between humans and two versions of ChatGPT, GPT-3.5 and GPT-4, across various psycholinguistic dimensions (Figure \ref{overview}a). Based on categories explored in \cite{lynott2020lancaster} and \cite{scott2019glasgow}, we evaluated the correspondence between humans and models along the following domains: (1) emotion, (2) salience, (3) mental visualization, (5) sensory, and (6) motor. Each domain consists of several dimensions (see Table \ref{definition} for definitions for each dimension). These dimensions provide comprehensive coverage for understanding the spectrum of human lexical-conceptual processing explored in previous studies (e.g., \cite{warriner2013norms, pereira2018toward, connell2018interoception}), from socio-emotional aspects, abstract mental imagery, to direct bodily experience (Figure \ref{overview}b). To examine the extent to which language can shape human-like conceptual representations, we investigated the alignment between word ratings from LLMs and human assessments for 4,442 words spanning the various psycholinguistic dimensions. To explore the role of sensory experience in conceptual formation, we analyzed how GPT-4's exposure to visual information may affect its representation of concepts compared to GPT-3.5. Finally, we validated LLMs' responses to ensure the robustness of our results. We discerned two key findings: firstly, both LLM versions closely mirrored human ratings in abstract domains, yet their performance in sensory and motor domains was less aligned, with GPT-4 showing superior performance over GPT-3.5. Secondly, the enhancements observed in GPT-4 are linked to its visual training, which seemingly extends benefits to related domains such as haptics and imageability. These results establish that language alone can support partial but not fully human-level conceptual representations, and that extending experience even into a single sensory domain can enhance model performance across domains.

\begin{table}[!ht]
\caption{Definitions of each dimension}
\centering
\begin{tabularx}{\linewidth}{ll p{4cm} X}
\hline
Norms & Domain & Dimension & Definition\\
\hline
Glasgow & Emotion & Valence & Value or worth; representing something considered good or bad.
\\
   & & Dominance & The degree of control a word makes you feel.
\\
\\
  & Salience & Arousal & Excitement versus calmness.
\\
   & & Size & Dimensions, magnitude, or extent of an object or concept that a word refers to\\
   & & Gender & How strongly its meaning is associated with male or female behaviour. \\
\\
  & Mental visualization & Concreteness & A measure of how concrete or abstract something is. \\
   & & Imageability & How easy or difficult something is to imagine. \\
\\
Lancaster & Perceptual & Haptic, Auditory, Olfactory, Interoceptive, Visual, Gustatory & How much do you experience everyday concepts using six different perceptual senses\\
\\
   & Motor & Foot/leg, Hand/arm, Torso, Head excluding mouth, Mouth/throat & How much do you experience everyday concepts using actions from five different parts of the body.\\

\hline
\end{tabularx}
\label{definition}
\end{table}

\section{Results}
We collected word ratings from ChatGPT and compared them with ratings generated by humans from the Glasgow \cite{scott2019glasgow} and Lancaster norms \cite{lynott2020lancaster}. The set of words being rated comprises 4,442 words and is shared between the Glasgow and Lancaster norms. The model prompt and design (Figure \ref{overview}c) for ChatGPT was standardized to match the instructions given to human subjects, maintaining consistency with human-subject data collection. GPT-3.5 and GPT-4 were separately ran for four rounds to ensure reliability. Detailed information on the agreement between these rounds can be found in the supplementary material. With ChatGPT responses obtained, we assessed the model's overall similarity to human word ratings by calculating Spearman correlations for each dimension at both aggregate and individual levels (Sections 2.1). Next, we explored if GPT-4's additional visual training impacted its performance compared to GPT-3.5 (Section 2.2). To substantiate the correlations found between human and model ratings, we performed secondary analyses (Section 2.3). Considering debates regarding the distinct roles of grounding in concrete versus abstract concepts \cite{sakreida2013abstract, fini2021abstract}, we separately analyzed human-model correlations for concrete and abstract terms. To ensure LLMs' validity as cognitive models \cite{shapira2023clever}, we adopted standard validation techniques from human-subject research \cite{scott2019glasgow} for dimensions with notable human-model agreement.

\subsection{ChatGPTs strongly align with human representations in non-sensorimotor domains but diverge in sensorimotor domains}
\begin{figure}[h]
    \centering
    \includegraphics[width=\textwidth]{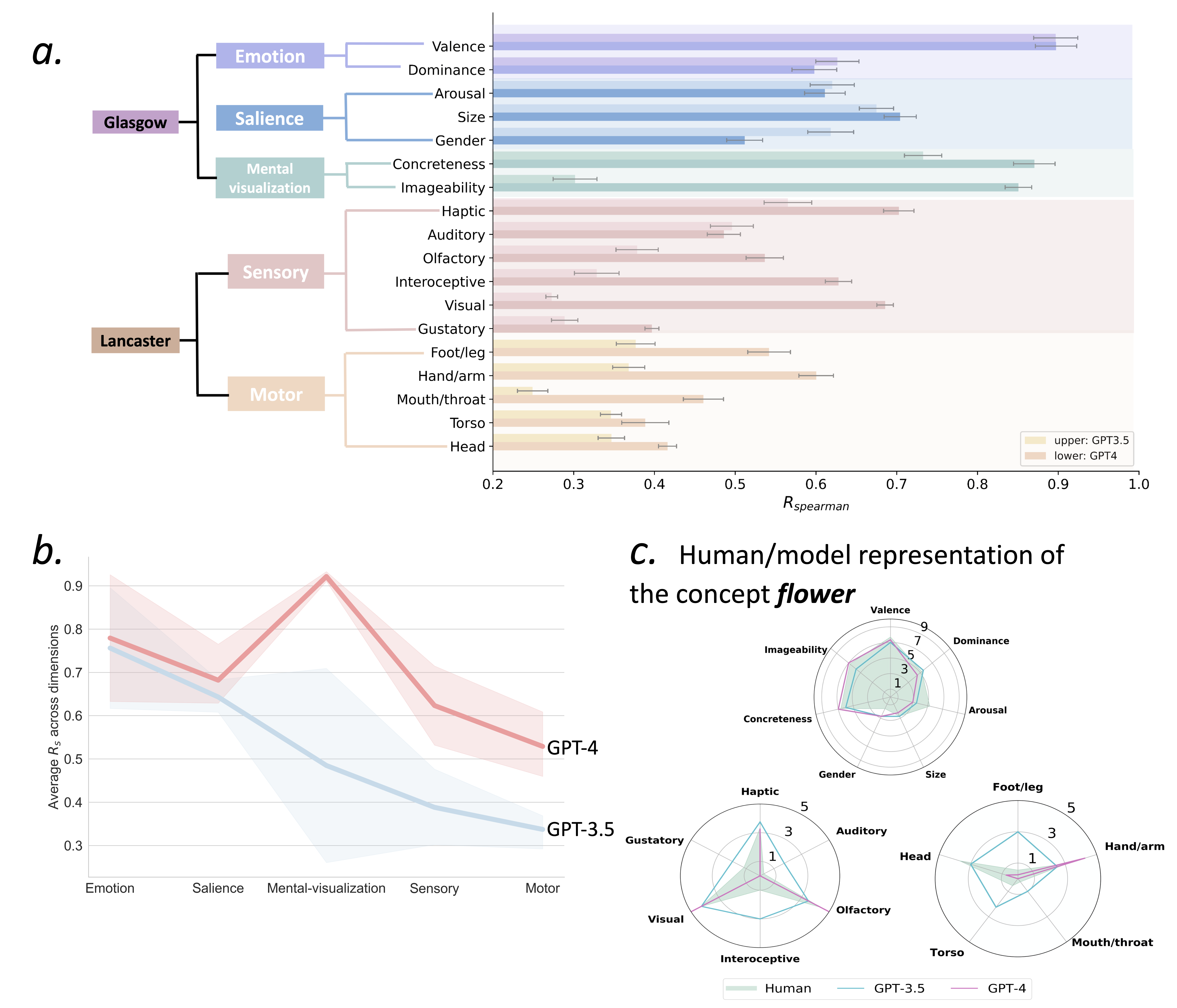}
    \caption{\textbf{a}. Spearman correlations between the aggregated ratings generated by ChatGPT (GPT-3.5 or GPT-4) and human ratings (x-axis) for each of the dimensions (y-axis). Error bars depict the 95\% confidence intervals, established through bootstrap resampling of 1,000 samples, comparing word ratings from both humans and ChatGPTs. \textbf{b}. Spearman correlations are presented for each domain (i.e., salience, emotion, etc.) by averaging across dimensions within the domain, with error bands indicating the 95\% confidence interval across the dimensions within the domain.  Human-model correlations are weaker in the sensory and motor categories than the emotion, salience, and visual categories. Additionally, GPT-4 consistently demonstrates stronger correlations with human ratings than GPT-3.5. \textbf{c}. An example of the representation of the concept \textit{flower} for humans, GPT-3.5, and GPT-4 in the dimensions within emotion, salience, and mental visualization categories (top), the sensory dimensions (bottom left), and the motor dimensions (bottom right). This illustrates that embodied experience is not required to comprehend the socio-emotional and abstract mental imagery associated with the concept \textit{flower}. Yet, the lack of embodied experience can result in significant differences from human representations in the sensory and motor facets of \textit{flower}.}
    \label{Aggregated}
\end{figure}

\paragraph{Aggregate analysis}
To evaluate the model's overall similarity to human word ratings, we calculated the Spearman rank correlation between the aggregated model-generated and human-generated rating matrices for each dimension. The model-generated ratings of each word were aggregated by averaging across the four rounds of GPT-3.5/GPT-4, and human-generated ratings were averaged across individual subjects (Figure \ref{Aggregated}).

\begin{figure}[h]
    \centering
    \includegraphics[width=0.9\textwidth]{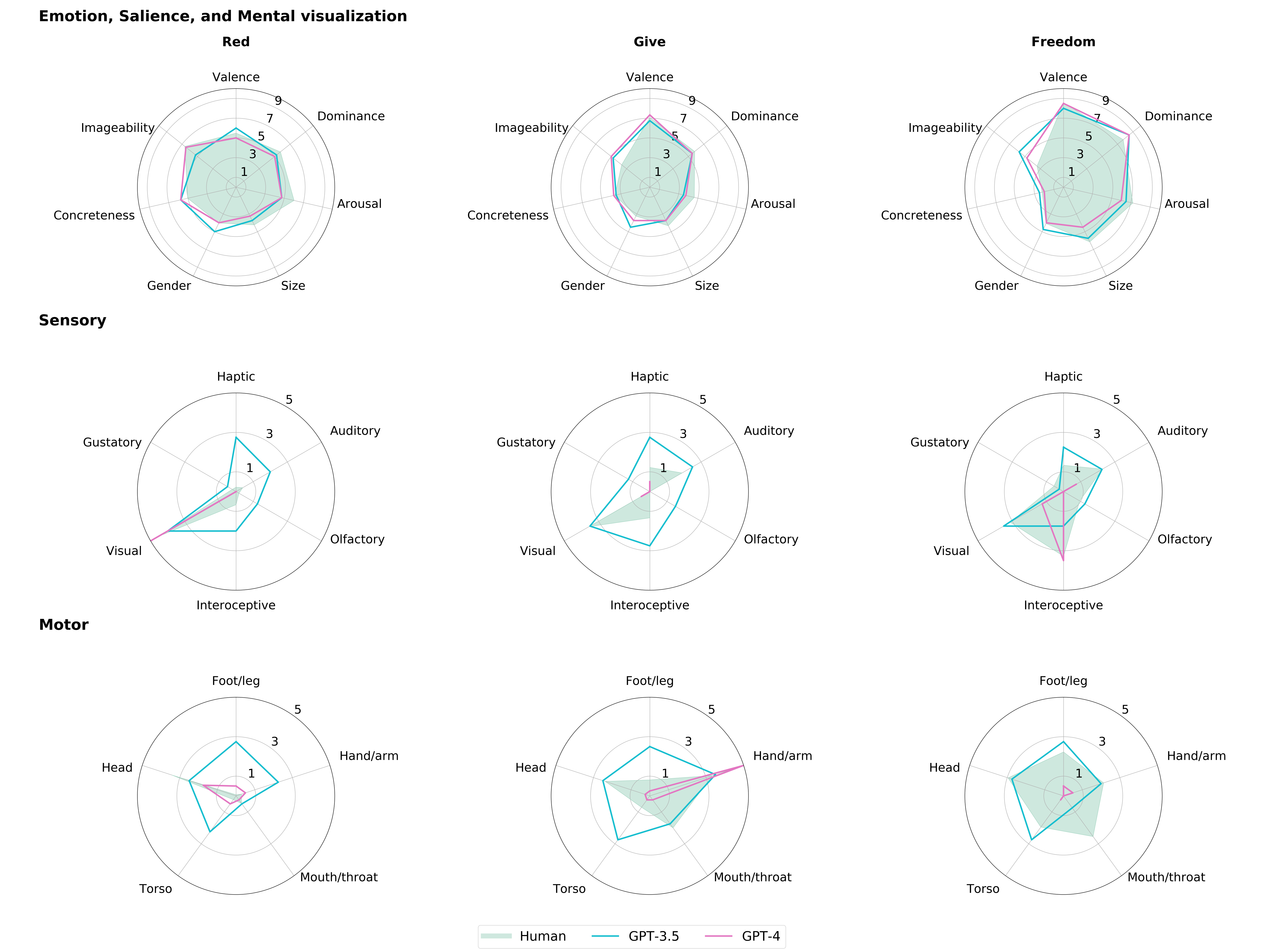}
    \caption{Radar plots for three individual concepts: red (color word), give (action verb), and freedom (abstract word). The numbers along the radial axis denote the aggregated ratings of human, GPT-3.5, and GPT-4 on each dimension. These examples highlight considerable alignment between the models and human judgments in the domains of emotion, salience, and mental visualization. Nonetheless, the charts also reveal significant divergences in the sensory and motor domains between the models' and humans' conceptual representations.}
    \label{Example}
\end{figure}

When comparing across different dimensions, both ChatGPT models exhibit strong correlations with human ratings in the dimensions of emotion, salience, and mental visualization ($Median = 0.64$ for GPT-3.5, $0.77$ for GPT-4), but significantly weaker correlations in the sensory and motor dimensions ($Median = 0.35$ for GPT-3.5 and $0.57$ for GPT-4), as suggested by Mann-Whitney U test, $U = 10.00, p = .004$ for human-GPT-3.5 correlations, and $U =12.00, p = .008$ for human-GPT-4 correlations. For example, both GPT-3.5 and GPT-4 show strong correlations with human ratings on valence, exceeding $0.90$, while in motor dimensions like foot/leg, the correlations are lower, at $0.38$ for GPT-3.5 and $0.58$ for GPT-4. When comparing between GPT-3.5 and GPT-4, GPT-4 ($Median = 0.64$) had a significantly greater correlations with human participants than GPT-3.5 ($Median = 0.38$), as suggested by a Wilcoxon signed-rank test, $W = 1.00, p < .001$. For instance, the imageability dimension reveals a substantial difference, with a human-GPT-3.5 correlation of $0.26$ compared to a human-GPT-4 correlation of $0.91$, and in the visual dimension, a human-GPT-3.5 correlation of $0.27$ compared with a human-GPT-4 correlation of $0.76$. Figure \ref{Example} presents illustrations of how models and humans represent three distinct concepts.

\begin{figure}[h] 
    \centering
    \includegraphics[width=\textwidth]{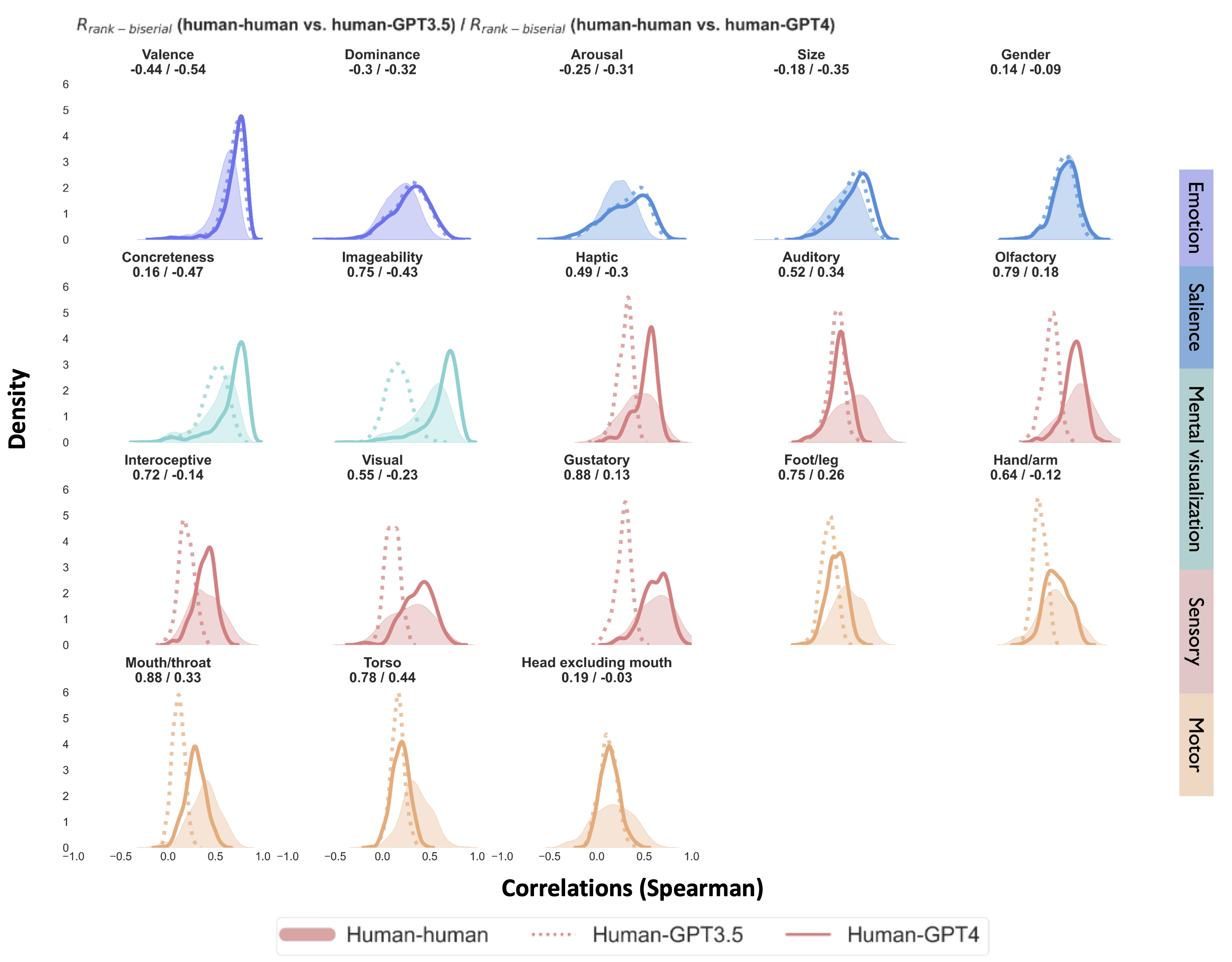}
    \caption{Individual-level correlations comparing ratings for pairs of humans (human-human), humans and GPT-3.5 (human-GPT3.5), and humans and GPT-4 (human-GPT4). The rank-biserial effect size values were reported in each dimension, where the value before "/" denotes the size between human-human and human-GPT3.5 and the value after "/" denotes between human-human and human-GPT4. A value close to -1 or 1 separately indicates values in the first group (human-human) are generally smaller or larger than the second group (human-GPT).}
    \label{Individual}
\end{figure}

\paragraph{Individual analysis}
To examine the similarity between human and model conceptual representations while considering individual variability, we undertook an individual-level analysis. We constructed pairwise Spearman correlations for every possible pairing in three distinct scenarios: a) between each pair of individual human participants (human-human), b) between each human participant and an individual instance of GPT-3.5 (human-GPT3.5), and c) between each human participant and an individual instance of GPT-4 (human-GPT4). This resulted in three distributions comprising human-human, human-GPT3.5, and human-GPT4 pairwise correlations, as depicted in Figure \ref{Individual}. This comparison highlights the degree of similarity between individual model runs and human participants, using the human-human correlations as a benchmark for inter-person reliability. To quantify the standardized distance between the human-human and model-human correlation distributions, we employed the rank-biserial correlation ($r_{rb}$), a common measure of effect size for non-parametric tests. A value of 0 signifies that the two samples come from identical populations, while values close to -1 or 1 indicate that values in one group (human-human) are generally lower or higher than those in the second group (human-GPT3.5/4), respectively. To facilitate result presentation and effect size interpretation, we adhered to the commonly accepted criteria for $r_{rb}$ \cite{cohen2013statistical}; in particular, an effect size between -0.10 and 0.10 is considered negligible. 

The individual-level findings are consistent with the aggregated findings, showing stronger model similarities with humans in the domains of emotion, salience, and mental visualization, and weaker similarities in sensory and motor domains. Meanwhile, human-GPT4 outperformed human-GPT3.5 in multiple dimensions. In the domains of emotion (valence and dominance) and salience (arousal and size), with gender being an exception, both human-GPT3.5 and human-GPT4 correlations were equal to or stronger than human-human correlations  ($r_{rb} < 0.10$), suggesting that a single run from the models could match or surpass the reliability of one human compared to another. In some dimensions, such as mental visualization (concreteness and imageability), four sensory dimensions (haptic, interoceptive, visual, and gustatory), and one motor dimension (hand/arm), GPT-4 but not GPT-3.5 showed strong correlations with individual human ratings, aligning as closely as humans do with one another ($r_{rb} > 0.10$). However, for other sensory and motor dimensions, neither GPT-3.5 nor GPT-4 achieved correlations on par with human-human benchmarks ($r_{rb} > 0.10$), indicating a divergence from individual-level human-like conceptual representations in these areas. 

\subsection{The improved alignment of GPT-4 with human representations is associated with its visual learning}
Given that GPT-4 showed larger correlations with human ratings compared to GPT-3.5, and that it underwent additional visual training with image inputs\cite{openai2023gpt4}, we sought to explore if this visual training contributed to its improved performance. However, isolating the impact of visual training is challenging due to the limited public information on how GPT-4 differs from GPT-3.5, beyond the additional visual domain training. To address this, we piloted an analysis characterizing a potential association between GPT-4's improvement over GPT-3.5 and its added visual input. The rationale underlying this analysis is that if the added visual training in GPT-4 contributes to its improved correlations with human ratings compared to GPT-3.5, this effect should be particularly noticeable in dimensions closely related to visual processing. Specifically, we would expect to see marked improvements in the visual dimension itself, as well as in dimensions such as imageability, which have been identified as having visual components in prior research \cite{scott2019glasgow}.

\begin{figure}[h]
    \centering
    \includegraphics[width=\textwidth]{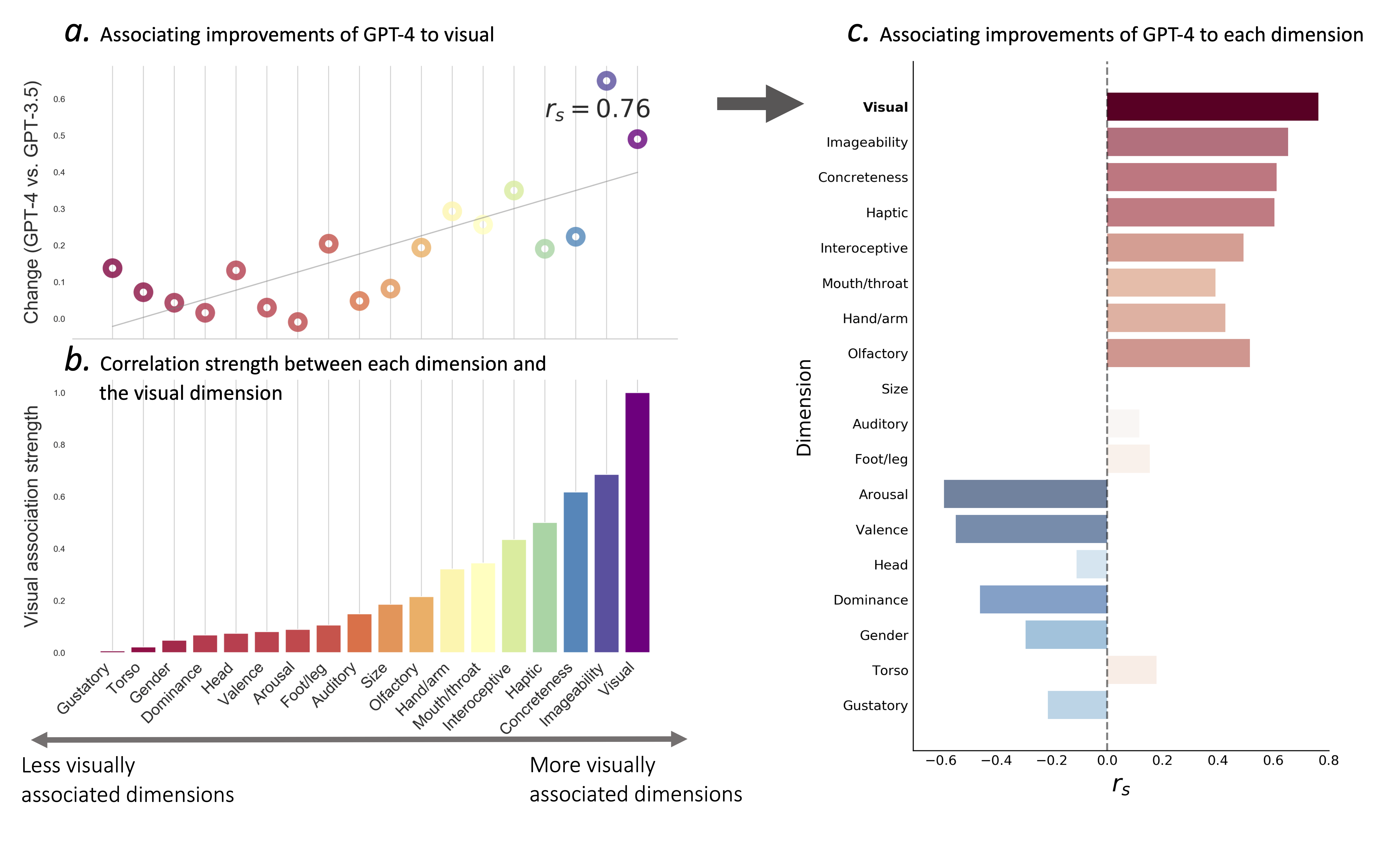}
    \caption{\textbf{a} The change in correlations between GPT-4 and human ratings compared to those of GPT-3.5 for each dimension (y-axis) strongly correlates with the correlation between ratings for each dimension and for the visual dimension (x-axis, $r_s = 0.76$). Dimensions more strongly associated with the visual dimension, such as imageability and interoceptive, exhibited greater improvement from GPT-3.5 to GPT-4. \textbf{b}  The absolute values of the Spearman correlation coefficients, based on human ratings \cite{lynott2020lancaster, scott2019glasgow}, reflect the association strength of each dimension with the visual domain. A higher coefficient signifies a stronger link to visual processing. \textbf{c} Associating improvements in GPT-4 to all dimensions, including non-visual dimensions using the same approach as in \textbf{a} and \textbf{b}, but with different reference dimensions instead of vision. The results show that the association with performance improvement is most evident when considering the visual dimension as the reference. As the reference shifts further away from the visual dimension, such association becomes increasingly weaker or even negative.}
    \label{Visual}
\end{figure}

To empirically evaluate the interconnectedness of psycholinguistic dimensions, we anchored our analysis on the visual dimension, using it as a reference point. We then determined the strength of its relationship with other dimensions by computing the absolute Spearman correlation coefficients, grounded in human rating data as reported in \cite{lynott2020lancaster, scott2019glasgow}. Higher absolute correlation coefficients indicate a stronger association with the visual dimension. For example, as illustrated in Figure \ref{Visual}b, dimensions such as concreteness and imageability are strongly associated with the visual dimension, whereas dimensions like gustatory and torso show minimal visual association. To evaluate the extent of improvement in GPT-4 over GPT-3.5, we calculated the difference between human-GPT4 and human-GPT3.5 correlations for each dimension separately. A higher value on a specific dimension indicates a greater improvement in GPT-4's correlation with human data compared to GPT-3.5 on that dimension (Figure \ref{Visual}a, y-axis).

We observed a strong pattern, where dimensions most strongly associated with the visual domain showed more substantial improvements in GPT-4 compared to GPT-3.5 (Figure \ref{Visual}a). This observation was supported by a large correlation between the degree of GPT-4's improvement over GPT-3.5 in a dimension and how visually associated that dimension is ($r_{s}(N=18) = 0.77, p < .001$), suggesting a correlation between visual-domain learning and the models' enhanced similarities with humans in visual-related processes. 

To rule out the possibility of analytical artifacts, we conducted the same analysis but shifted our reference from the visual dimension to other non-visual dimensions. Specifically, we examined associations with non-visual dimensions to ascertain whether the \textit{dimension-specific change} pattern, as depicted in \ref{Visual}a, persisted. If the visual aspect is indeed a key and unique factor in GPT-4's performance improvement, weaker patterns would be anticipated when non-visual dimensions serve as our reference. The findings showed that the \textit{dimension-specific change} is most pronounced when utilizing the visual dimension as the reference dimension. As the reference dimension becomes more unrelated to the visual, the \textit{dimension-specific change} pattern correspondingly weakens. In the case of dimensions with minimal or no association with the visual, we observed null or even negative patterns. These results reinforce the hypothesis that the transition from GPT-3.5 to GPT-4 may be attributed to GPT-4's visual domain learning.

\subsection{Secondary analyses for validating the results}
\begin{figure}[h]
    \centering
    \includegraphics[width=1\textwidth]{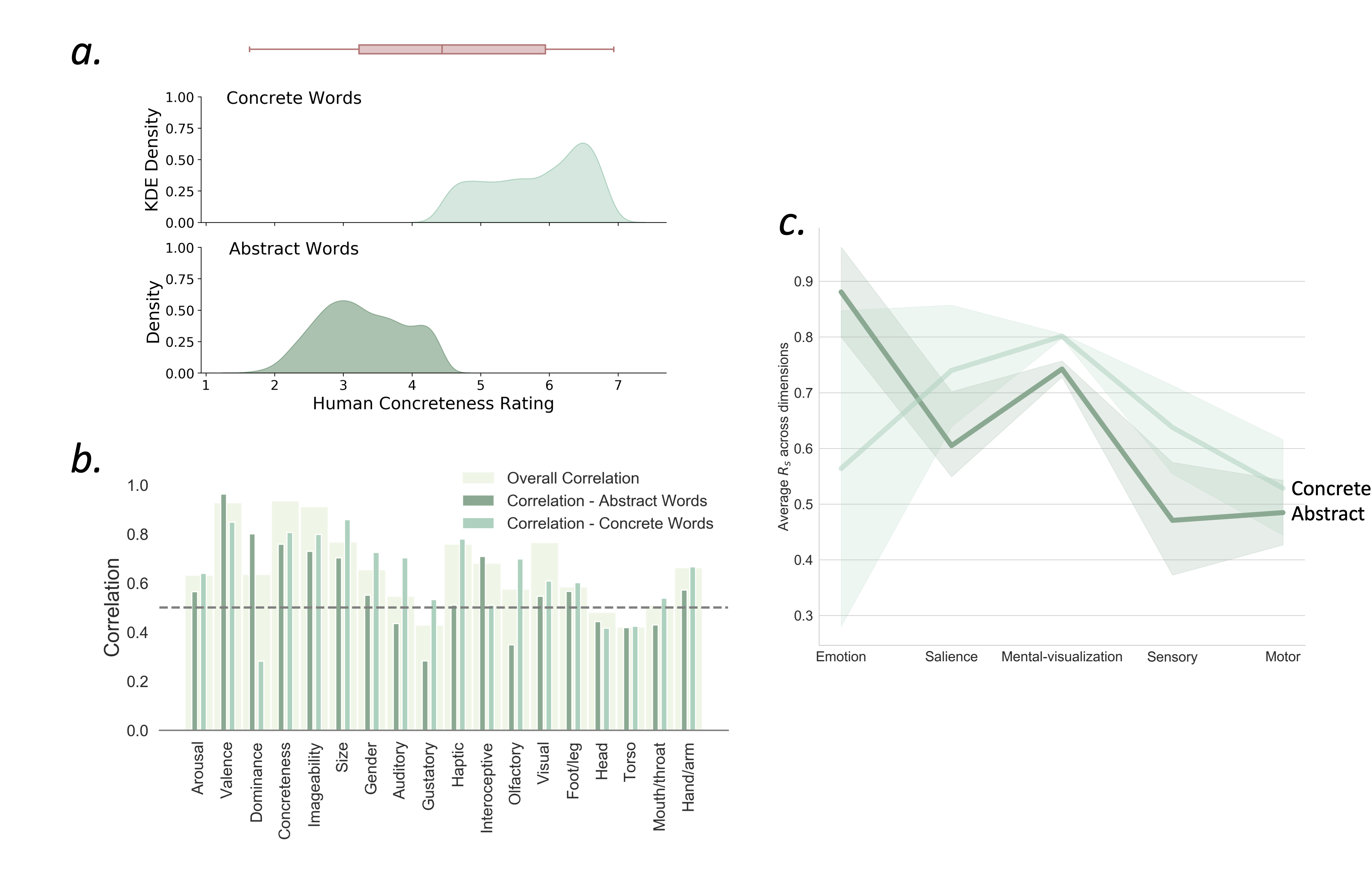}
    \caption{\textbf{a}. Using concreteness ratings from the Glasgow norm, words were classified into two categories: \textit{abstract words}, where the concreteness ratings fall below the median, and \textit{concrete words}, where the ratings are above the median. \textbf{b}. Correlations between the aggregated GPT-4 and human ratings, separately for \textit{concrete} and \textit{abstract} words. The background green bars represent the overall correlations between the aggregated GPT-4 and human ratings, without distinguishing between \textit{concrete} and \textit{abstract} words, identical to the one presented in Figure \ref{Aggregated}. \textbf{c}. Spearman correlations for \textit{concrete} and \textit{abstract} words are presented for each domain (e.g., salience, emotion). These are averaged across dimensions within each domain, and the error bands represent a 95\% confidence interval. Generally, the correlations of human ratings with ChatGPT for both \textit{concrete} and \textit{abstract} words align with the overall findings in Section 2.1.} 
    \label{concreteness}
\end{figure}

\paragraph{Concrete and abstract words} We examined the human-model correlations for concrete and abstract words, given their possible distinct grounding processes discussed in previous literature\cite{sakreida2013abstract, fini2021abstract}. Using concreteness ratings from the Glasgow norm \cite{scott2019glasgow}, we classified words into two categories: \textit{abstract words}, where the concreteness ratings fall below the median, and \textit{concrete words}, where the ratings are above the median (Figure \ref{concreteness}a). We subsequently analyzed Spearman correlations between the aggregated ChatGPTs and human ratings separately for these \textit{concrete} and \textit{abstract} words. 

The correlations of human ratings with ChatGPT for both \textit{concrete} and \textit{abstract} words align with findings in Section 2.1, as indicated by Mann-Whitney U tests. For both word types, ChatGPT models showed strong correlations in emotion, salience, and mental visualization dimensions, but siganificantly weaker in sensory and motor dimensions (GPT-3.5: $U = 64.0, N_1 = 7, N_2 = 11, p = .020$ for \textit{abstract} words, and $U = 67.0, p = .008$ for \textit{concrete} words; GPT-4: $U =70.0, p = .003$ for \textit{abstract} words, and $U = 61.0, p = .044$ for \textit{concrete} words). GPT-4 outperformed Chat3.5 for both \textit{abstract} ($W = 2.0, N = 18, p < .001$) and \textit{concrete} words ($W = 9.0, p < .001$). Additionally, Wilcoxon signed-rank tests suggested that human-model correlations for \textit{abstract} words ($Median_{GPT-3.5} = 0.35, Median_{GPT-4} = 0.56$) were significantly lower than for \textit{concrete} words ($Median_{GPT-3.5} = 0.43, Median_{GPT-4} = 0.65$), with $W = 128.0, N = 18, p = .033$ for both models. Figure \ref{concreteness}(b-c) illustrates human-GPT4 correlations for \textit{concrete} and \textit{abstract} words separately. 

\paragraph{Validating ChatGPT responses}
Recognizing the critical need for validity in LLM applications \cite{shapira2023clever, ma2023tomchallenges, frank2023large}, as emphasized earlier, we adhered to established human test validation methods \cite{lynott2020lancaster, scott2019glasgow}, evaluating ChatGPT (GPT-3.5 and GPT-4) against a set of alternate norms that mimic the Glasgow and Lancaster measures. Consistent responses across these varied norms would suggest that ChatGPT's outputs are not just plausible but are validated against human standards.

For the Glasgow norm, validation norms include dimensions of valence, arousal, and dominance from \cite{warriner2013norms}, and imageability \cite{cortese2004imageability} and concreteness from\cite{brysbaert2014concreteness}. For the Lancaster norm, which lacks directly comparable validation norms, we included dimensions that are conceptually similar, such as taste and grasp \cite{amsel2012perceptual}. In human ratings, taste is expected to strongly correlate with the gustatory dimension in Lancaster, while grasp shows moderate correlations with the hand/arm and haptic dimensions. We selected these validation norms for several reasons: a) they are publicly accessible; b) they have been widely used in human-subject studies, lending to their credibility; and c) they cover dimensions in either the Glasgow or Lancaster norms where GPT-3.5 or GPT-4 show strong correlations (i.e., $r_{s} > 0.6$) with human data.

As detailed in Table \ref{validation}, we first evaluated ChatGPT's responses on the validation norms, then computed Spearman correlations between humans and ChatGPT for these norms. Subsequently, we calculated correlations for ChatGPT ratings between the original Glasgow/Lancaster norms and the validation norms. The results revealed that human-ChatGPT correlations from the validation norms closely resemble those from the Glasgow/Lancaster norms. For instance, the correlation between human ratings and GPT-3.5 on valence was 0.83 in the validation norm, compared to 0.90 in the Glasgow norm. Moreover, the correlation strength of ChatGPT ratings between the validation norms and the Glasgow/Lancaster norms is as high as the correlation strength of human ratings across these norm sets. For instance, the correlation for GPT-4 ratings on the hand/arm dimension between the validation and the Lancaster norms was 0.68, compared to the 0.55 correlation of human ratings across these norms. Converging results based on an independent set of norms lend support to the validity of our conclusions derived from the ChatGPT ratings.

\begin{table}[ht] 
\centering
\begin{tabular}{llcccc}
\toprule
Model & Dimension & \multicolumn{2}{c}{GPT-human} & \multicolumn{2}{c}{Original-validation} \\
& & \multicolumn{1}{c}{Original} & \multicolumn{1}{c}{Validation} & \multicolumn{1}{c}{GPT} & \multicolumn{1}{c}{Human}\\
\midrule
GPT-3.5 & Valence & 0.90 & 0.83 & 0.90 & 0.93 \\
     & Dominance & 0.62 & 0.67 & 0.82 & 0.69 \\ 
    & Arousal & 0.64 & 0.47 & 0.55 & 0.62\\
    & Concreteness & 0.71 & 0.63 & 0.61 & 0.93\\
\midrule
GPT-4 & Valence & 0.93 & 0.88 & 0.92 & 0.93\\
    & Dominance & 0.63 & 0.67 & 0.86 & 0.69\\
    & Arousal & 0.63 & 0.43 & 0.54 & 0.62\\
    & Concreteness & 0.93 & 0.87 & 0.88 & 0.93\\
    & Imageability & 0.91 & 0.77 & 0.83 & 0.89\\
    & Haptic & 0.75 & 0.88 & 0.55 & 0.55\\
    & Hand/arm & 0.66 & 0.88 & 0.68 & 0.55\\
\bottomrule
\end{tabular}
\caption{Spearman correlations between ChatGPT and humans (\textit{GPT-human}) ratings separately for the Glasgow/Lancaster norm (\textit{original}) and the validation norms (\textit{validation}), and correlations between the Glasgow/Lancaster norm and the validation norms (\textit{Original-validation}) separately for GPT ratings (\textit{GPT}) and human ratings (\textit{Human}).}
\label{validation}
\end{table}

\section{Discussion}
In this study, we used large language models (LLMs) to test the limits of conceptual knowledge acquisition by quantifying what structures of human conceptual knowledge can be learned from limited inputs—namely, from language alone, or language and vision. We showed that some domains, such as emotion, salience, and mental visualization may not rely heavily on sensory grounding and can be obtained through language alone. However, we observed a noticeable disparity between humans and LLMs in sensorimotor domains; that is, learning from language alone yields impoverished sensorimotor knowledge. In light of our findings and their relevance to the ongoing debate about the necessity of embodied grounding for achieving human-level conceptual representation \cite{phildeeplearning2023, chemero2023llms}, these findings suggest that while some aspects of conceptual representations may be detached from sensory experience, a considerable degree of sensory input appears essential. Take the concept of \textit{red} for instance. Language alone can capture certain conceptual connotations of the color "red" insofar as they emerge from relationships among words in context. However, \textit{red} is more than a color word; the sensory experience of redness may cut across linguistic contexts and may implicitly shape our conceptual knowledge even when the word "red" does reach the threshold of linguistic articulation, so as to form diverse relationships across objects and experiences in the world around us. From the crispness of an apple to the urgency of a stop sign, 'red' binds these disparate elements into a coherent category. This kind of associative perceptual learning, where 'red' becomes a nexus of interconnected meanings and sensations, is difficult to achieve through language alone. These aspects are likely ingrained in human conceptual knowledge through real-world interactions, underscoring the potential necessity of embodied experiences for comprehensive sensory perception, physical action, and perceptual understanding. 

The current study exemplifies the potential benefits of multimodal learning that 'the whole is greater than the sum of its parts', showing how the integration of diverse modalities of inputs leads to a more human-like understanding than what each modality could offer independently. We observed a strong association between GPT-4's advancements and its augmented visual learning. These improvements extend beyond the visual dimension, positively impacting related dimensions like haptics and imageability. This suggests that visual learning may play an important role in aligning GPT-4's performance more closely with human behaviors. Taking this argument further, one could imagine if GPT-4 had incorporated more sensorimotor components, their performance would become even more aligned with human behavior. However, our findings indicate that there may not be a linear relationship between the amount of sensorimotor components and LLM's human-like performance. Instead, our findings indicate that a potential for knowledge to transfer between related domains in LLMs: learning in one modality can enhance understanding in other neighboring modalities. This transferable representation is well observed in humans \cite{peelen2014nonvisual, noppeney2003effects}. For instance, humans can acquire object-shape knowledge through both visual and tactile experiences \cite{peelen2014nonvisual}. Given the architecture and learning mechanisms of GPT-4, where representations are encoded in a continuous, high-dimensional embedding space, inputs from multiple modalities may fuse or shift embeddings in this continuous space. The smooth, continuous structure of this embedding space may underlie our observation that knowledge derived from one modality seems to spread across other related modalities \cite{shepard1987toward, landauer1997solution, mikolov2013linguistic}. Additionally, it points to a possibility that to achieve human-like conceptual representations, full sensorimotor access might not be necessary; partial access could suffice to span the breadth of human experience. Future research should explore the extent of sensory access needed and the limits of knowledge transfer across domains in multimodal models.The progression of LLMs towards integrating additional modalities—as seen in multimodal speech and text processing in Whisper \cite{radford2022whisper} and embodied vision-language-action models like RT-2 \cite{rt22023arxiv}—opens exciting prospects for further understanding and harnessing the potential of multimodal learning.

Our findings align with the dual-coding knowledge theory \cite{bi2021dual, ralph2017neural}, positing the existence of language/cognition-derived knowledge independent of sensory experience. However, our findings extend this theory: through analyzing a broad range of concepts and precisely separating language from other knowledge sources, we provide further insights into the extent to which language alone can provide basis for shaping complex concepts and the necessity of embodied experiences. Our results also hint at the potential for knowledge transfer across different domains, an aspect not extensively explored in prior research with congenitally blind population. We note, however, that although LLMs seem to implicitly approximate some bodily knowledge, they obtain this by consuming vast amounts of text, orders of magnitude larger than the volume of language a human is exposed to in their entire lifetime. This suggests that, while in the limit multimodal knowledge can be synthesized from language alone, this kind of learning is inefficient. In contrast, human learning is inherently multimodal and multisensory from the outset. Infants, for instance, absorb knowledge through a variety of interactive sensorimotor channels in social contexts and interactions, which might explain their remarkable efficiency and effectiveness as learners \cite{yu2016social}.

One critical question arises regarding the extent to which the word rating tasks in our study accurately reflect genuine conceptual representations in naturalistic settings. We maintain that well-validated psycholinguistic norms offer an essential glimpse into key facets of human cognition, a standpoint supported by previous research \cite{winter2019sensory, winter2017words, perlman2018iconicity, winter2016taste}. Although these norms might not align precisely with the intricacies of complex reasoning and language usage, they are fundamental in outlining our cognitive functions. In practical language applications, such as communication, writing, metaphorical expressions, and iconicity, sensorimotor representations of concepts may play a larger role than in simplified rating tasks. Nonetheless, quantifying the interrelations between conceptual representations in humans and models in real-world contexts remains a formidable task. To meet this theoretical challenge, future research should strive to operationalize and test constructs that can accommodate richer, communicative contexts. Our individual-level analyses also reveal a substantial variability in human ratings (Figure \ref{Individual}). Each LLM, however, essentially represents a singular `participant', reflecting aggregated language patterns and inputs from text generated by millions of humans. In essence, while LLMs may capture a central tendency, they do not necessarily express the diversity and idiosyncrasies characteristic of individual humans \cite{dillion2023can}. Novel learning mechanisms and individualized training regimes will be needed to more precisely model the unique tapestry of conceptual understanding learned by individual humans.

\section{Methods}
\label{sec:headings}
\paragraph{Psycholinguistic Norms}
We used the Glasgow Norms \cite{scott2019glasgow} and the Lancaster Sensorimotor Norms (henceforth the Lancaster Norms; \cite{lynott2020lancaster}) as human psycholinguistic word rating norms (see Table \ref{definition} for their  dimensions). Taken together, the two norms offer comprehensive coverage of the included dimensions, both of which cover a large number of words. 

The Glasgow Norms consist of normative ratings for 5,553 English words across nine dimensions \cite{scott2019glasgow}. We selected the Glasgow Norms due to its large-scale data and highly standardized data collection process: the same participants rated all dimensions for any given subset of words, with an average of 33 participants per word. The nine dimensions include emotional arousal, valence, dominance, concreteness, imageability, size, gender association, familiarity, and age of acquisition. In our study, we excluded familiarity and age of acquisition, as familiarity is less dependent on semantic and conceptual representation \cite{balota2001subjective} and therefore less relevant to our research focus, while age of acquisition is neither central to our focus and nor a valid question for LLMs to answer. The validity of the Glasgow Norms has been demonstrated through strong correlations with 18 different sets of other psycholinguistic norms. Scott et al. \cite{scott2019glasgow} conducted principal component analyses and identified three main categories underlying these dimensions: emotion (valence and dominance), salience (arousal, size, and gender), and mental visualization (concreteness and imageability). We adopt their validated structure for categorizing those dimensions.

The Lancaster Norms present multidimensional measures encompassing sensory and motor strengths for approximately 40,000 English words \cite{lynott2020lancaster}. These norms include six sensory dimensions (haptic, auditory, olfactory, interoceptive, visual, gustatory) and five motor dimensions (foot/leg, hand/arm, mouth/throat, torso, head excluding mouth). The sensorimotor properties of words are considered highly embodied, as they require human raters to utilize their everyday perceptual senses and bodily experiences to gauge each word. Data were collected from 3,500 unique participants, with each participant rating on average 7.12 lists for either the sensory or motor dimensions. Each list comprised 58 words, including 48 target words, five control words, and five calibration words. The fixed sets of five control words were randomly interspersed to each item list to ensure the quality of participants' ratings and the five calibration words were presented at the beginning of each item list to introduce participants to unambiguous examples for rating. The Lancaster Norms were chosen primarily because they provide a detailed and comprehensive representation of a word's perceived sensorimotor strengths across 11 dimensions, covering all senses and the five most common action effectors. The norms exhibit high reliability, displaying substantial consistency across all dimensions, and their validity is demonstrated by their ability to accurately represent lexical decision-making behavior from two distinct databases \cite{lynott2020lancaster}. 

We adhered to the design of the human-subject data collection (Figure \ref{overview}b; \cite{lynott2020lancaster, scott2019glasgow}). For the Glasgow measures, the 5,553 words were divided into 40 lists, with eight lists containing 101 words per list and 32 lists containing 150 words per list. The models rated all words in a list for one dimension before moving on to the next dimension, and so forth. The order of words within each dimension and the order of dimensions within each testing round were randomized. For the Lancaster measures, there are in total 39,707 available words with cleaned and validated sensorimotor ratings. We first extracted 4,442 words overlapping with the 5,553 words in the Glasgow measures. Following the practice in the Lancaster Norms, we obtained the frequency and concreteness measures \cite{brysbaert2014concreteness} of these 4,442 words and attempted to perform quantile splits over them to generate item lists that maximally resemble those in the Lancaster Norms. However, since more than 95\% of the 4,442 words have a “percentage of being known” greater than 95\%, we considered the majority of these words to be recognizable by human raters. Thus, we did not perform a quantile split of these words over word frequency. We instead implemented a quantile split based on their concreteness ratings with four quantile bins in the intervals 1.19 to 2.46, 2.46 to 3.61, 3.61 to 4.57, 4.57 to 5.00. 

Next, we generated four sublists based on the concreteness rating quantile split and randomly selected 12 words from each sublist without replacement to create 48 words for each item list. We further appended the five calibration words (sensory dimensions: \textit{account}, \textit{breath}, \textit{echo},\textit{hungry}, \textit{liquid}; motor dimensions: \textit{shell}, \textit{tourism}, \textit{driving}, \textit{breathe}, \textit{listen}) to the beginning of each list. Finally, we randomly inserted five control words (sensory dimensions: \textit{grass}, \textit{honey}, \textit{laughing}, \textit{noisy}, \textit{republic};  motor dimensions: \textit{bite}, \textit{enduring}, \textit{moving}, \textit{stare}, \textit{vintage}) into these lists to form 93 complete items lists, each containing 58 words ready to be rated separately for sensory and motor dimensions. The order of words within each item list and the order of dimensions to rate for each round were randomized. For GPT-4, since rating 58 words occasionally exceeded the token limit, we divided these 93 lists to form 186 lists while keeping the implementation of calibration and control words and randomization of rating order consistent. 

\paragraph{Models}
The selection of parameters in our study was based on methodological considerations aimed at optimizing the accuracy and consistency of the model outputs. We employed the gpt-3.5-turbo-0301 and gpt-4-0314 models from the OpenAI API to evaluate GPT-3.5 and GPT-4, respectively. The temperature parameter was set to 0, following recommendations in \cite{binz2023using}) and \cite{kosinski2023theory}, to ensure deterministic, consistent responses without random variations. The maximum token length was set to 2,048, a decision informed by the need to capture complete responses without truncation. To enhance the reliability of our results, we implemented four rounds of testing for each model. This approach allowed us to cross-verify the consistency of the outputs across multiple iterations (see supplementary for the agreement between these rounds).

\paragraph{Testing procedure}
The model prompt to ChatGPT was kept identical to the instructions that human subjects received. However, we made minor adjustments to the prompt to ensure that the responses followed the expected format (e.g., word - rating) When given testing items from the Lancaster norms, the model consistently responded that it does not possess a biological body and therefore cannot experience the word through sensing or moving. To address this, we modified the instruction from "to what extent do you experience" to "to what extent do human beings experience", and we applied the same changes to the Glasgow norms for consistency. Although the LLM is responding on behalf of human experience, it is still utilizing its internal representations to provide answers. These representations are derived from extensive training on human-generated text, which makes the responses valid as they reflect the collective knowledge and understanding of humans.

Images or tables used in human-subject tasks were converted to text format. Moreover, the online rating portal of the Lancaster Norm used a graphic demonstration of the five body parts for the action-executing effector ratings. Because GPT-3.5 currently do not support such visual inputs in the prompts, we decided to describe these five body parts with words in the prompts for both GPT-3.5 and GPT-4 instead (see the supplementary material for a comparison between the instructions given to human subjects and the adapted version provided to the models).

\paragraph{Words being analyzed}
For more uniform comparisons across various dimensions, we included words common to both the Glasgow and Lancaster norms in our analysis, resulting in a total of 4,442 words. Each of these words has corresponding ratings across all evaluated dimensions.

\paragraph{Individual-level pairwise correlations}
For individual-level analysis, we computed pairwise Spearman correlations across every possible pairing within three scenarios: a) between each pair of human participants (human-human), b) between each human and a single instance of GPT-3.5 (human-GPT3.5), and c) between each human and a single instance of GPT-4 (human-GPT4).

In the human-human correlations, each participant evaluated only a subset of words. In the Glasgow norm, participants rated one of either eight lists (comprising 808 words in total, with 101 words per list) or 32 lists (from a pool of 4,800 words, with 150 words per list). Each list received ratings from 32-36 participants, and there was no overlap in words across different lists. Pairwise correlations were calculated within each list, and these were aggregated, resulting in a total of 22,730 pairs for constructing the overall distribution for each dimension in the Glasgow norm.

In the Lancaster norm, the sensory component involved 2,625 participants (averaging 5.99 lists each) and the motor component had 1,933 participants (averaging 8.67 lists each). Each list included 48 test items, along with a constant set of five calibrator and five control words, totaling 58 items per list. Given the larger pool of 40,000 words in the Lancaster norm, the subset of 4,442 words resulted in some participants rating few items. To maintain a sufficient sample size for correlation calculations, we iterated through pairs of participants and included those with ratings for over 50 common words. This approach yielded 105 pairs for every sensory dimension and 196 pairs for every motor dimension, from which we constructed the correlation distributions.

In the human-GPT3.5 and human-GPT4 correlations, as each model iteration rated all 4,442 words, we generated pairs by matching each model run (out of four total runs) with individual human subjects across different lists. This approach yielded 5,476 pairs for the Glasgow norm. For the Lancaster norm, we paired humans and models based on having ratings for over 50 common words, mirroring the approach used in constructing human-human pairs. This process resulted in a total of 276 pairs for each sensory dimension and 376 pairs for each motor dimension, forming the basis for the correlation distributions.
\vspace{3em}
\paragraph{Acknowledgement}
We would like to express our sincere gratitude to James Magnuson and Sandeep Prasada, as well as to all members of the Brain, Language, and Cognition Lab, for their invaluable feedback. Additionally, we are indebted to Sara Sereno and Jack Taylor for their generous contribution in sharing the trial-level dataset of the Glasgow Norms. This research is supported by a grant from the Hong Kong Research Grants Council (Project \#PolyU15601520) and a Research Startup Fund from the Hong Kong Polytechnic University provided to Ping Li, and the Basque Government through the BERC 2022-2025 program and the Spanish State Research Agency through BCBL Severo Ochoa excellence accreditation CEX2020-001010/AEI/10.13039/501100011033 provided to Qihui Xu.

\bibliographystyle{unsrt}  
\bibliography{references} 





\end{document}


\title{Supplementary Material}
\date{}

\maketitle
\linenumbers

\section{Agreement across four model runs}
We employed intraclass correlation (ICC) to assess the consistency of responses across different runs of GPT-3.5 and GPT-4 for each dimension. ICC quantifies the degree of similarity or agreement between measurements made in groups or classes — in this case, the responses of different model runs. A high ICC value indicates strong agreement, while a low value suggests variability. The ICC results are detailed in Table \ref{tab:table_s1}. Although all coefficients are statistically significant, we have omitted significant values due to the large sample size. For GPT-3.5, we observed higher consistency in dimensions related to emotion, valence, and mental visualization, but lower consistency in sensory and motor domains. In comparison, GPT-4 demonstrated overall stronger agreement across runs than GPT-3.5, yet some motor dimensions still showed relatively weaker agreement. The findings indicate that in dimensions where model-human similarity is high (as detailed in the main text), the model demonstrates stable and consistent performance. Conversely, in dimensions with lower model-human similarity, the model's performance tends to be less consistent.
\begin{table}[h!]
\centering
\begin{tabular}{|llll|}
\hline
Domain & Dimension & GPT-3.5 & GPT-4 \\
\hline
Emotion & Valence & 0.93 & 0.95 \\
 & Dominance & 0.80 & 0.79\\
\hline
Valence & Arousal & 0.77 & 0.87 \\
 & Gender & 0.64 & 0.68 \\
 & Size & 0.78 & 0.82 \\
\hline
Meatal visualization & Imageability & 0.57 & 0.85 \\
 & Concreteness & 0.73 & 0.90\\
\hline
Sensory & Haptic & 0.39 & 0.80\\
& Auditory & 0.46 & 0.66\\
& Olfactory & 0.43 & 0.83\\
& Interoceptive & 0.39 & 0.65\\
& Visual & 0.28 & 0.72\\
& Gustatory & 0.55 & 0.89\\
\hline
Motor & Foot/leg & 0.26 & 0.59\\
 & Hand/arm & 0.22 & 0.67\\
 & Mouth/throat & 0.30 & 0.69\\
 & Torso & 0.20 & 0.44\\
 & Head & 0.23 & 0.36\\
\hline
\end{tabular}
\caption{Agreement across four model runs}
\label{tab:table_s1}
\end{table}

\section{Comparison between the instructions given to human subjects and the adapted version provided to the models}

\begin{longtable}{|>{\raggedright\arraybackslash}p{0.12\textwidth}|>{\raggedright\arraybackslash}p{0.44\textwidth}|>{\raggedright\arraybackslash}p{0.44\textwidth}|}
\caption{Comparison between the instructions given to human subjects and the adapted version provided to the models} \\
\hline
\textbf{Instruction type} & \textbf{Instruction - Humans} & \textbf{Instruction - GPT} \\ \hline
\endfirsthead 

\multicolumn{3}{c}{\tablename\ \thetable\ -- \textit{Continued from previous page}} \\
\hline
\textbf{Instruction type} & \textbf{Instruction - Humans} & \textbf{Instruction - GPT} \\ \hline
\endhead 

\multicolumn{2}{r}{\textit{Continued on next page}} \\
\endfoot 

\hline
\endlastfoot 

Glasgow (General instruction) & In this experiment, you will be rating a set of 101/150 words on \textbf{9 different scales – familiarity, concreteness, arousal, valence, dominance, imagaeability, size, gender, and age of acquisition}. These different scales assess different aspects of word meanings.
\vskip 0.05in
You will rate the entire word set on one scale, then rate them all again on the next scale, and so on. You will be given instructions about what each scale represents before you begin each scale.
\vskip 0.05in
Sometimes words have more than one meaning – for example, the word “nail” has one meaning related to fingers and one related to hammers In such cases, we will display the word in one of its meanings – “nail (finger)” or “nail” (hammer)” – or just by itself as “nail”. Please rate these words according to your first impression.
\vskip 0.05in
It is also possible that you may be presented with a word that you don’t know. If this happens, \textbf{there is a button “Unfamiliar word” located below the rating scale that you can click on to proceed to the next trial}.
\vskip 0.05in
Finally, it may sometimes be difficult to rate a word on a given scale. For example, the word “desk” might be difficult to categorise as either being a masculine or feminine thing. Likewise, the word “amusing” might be difficult to categorise as something that is big or small. When you are faced with such difficult decisions, please respond as best as you can without thinking too deeply – go with your intuitions. \textbf{Note: Click on the red information button in the upper right corner at any time to refer to the rating instructions again}. & In this experiment, you will be rating a set of 101/150 words on \textbf{eight different scales – familiarity, concreteness, arousal, valence, dominance, imagaeability, size, and gender}. These different scales assess different aspects of word meanings. 
\vskip 0.05in
You will rate the entire word set on one scale, then rate them all again on the next scale, and so on. You will be given instructions about what each scale represents before you begin each scale. 
\vskip 0.05in
Sometimes words have more than one meaning – for example, the word “nail” has one meaning related to fingers and one related to hammers. In such cases, we will display the word in one of its meanings – “nail (finger)” or “nail” (hammer)” – or just by itself as “nail”. Please rate these words according to your first impression. 
\vskip 0.05in
It is also possible that you may be presented with a word that you don’t know. If this happens, \textbf{please say “Unfamiliar word”}. 
\vskip 0.05in
Finally, it may sometimes be difficult to rate a word on a given scale. For example, the word “desk” might be difficult to categorise as either being a masculine or feminine thing. Likewise, the word “amusing” might be difficult to categorise as something that is big or small. When you are faced with such difficult decisions, please respond as best as you can without thinking too deeply – go with your intuitions.\\ \hline
Glasgow (Arousal)\footnote{Table reprinted from \cite{scott2019glasgow}. Copyright 2019 by Scott, G. G., Keitel, A., Becirspahic, M., Yao, B., \& Sereno, S. C. Reproduced with permission via the Creative Commons Attribution 4.0 International License (http:// creativecommons.org/licenses/by/4.0/).} & Arousal is a measure of excitement versus calmness. A word is AROUSING if it makes you feel stimulated, excited, frenzied, jittery, or wide-awake. A word is UNAROUSING if it makes you feel relaxed, calm, sluggish, dull, or sleepy.
 \vskip 0.05in
Please indicate how arousing \textbf{you think} each word is \textbf{on a scale of VERY UNAROUSING to VERY AROUSING, with the midpoint representing moderate arousal}. 
\includegraphics[width=\linewidth]{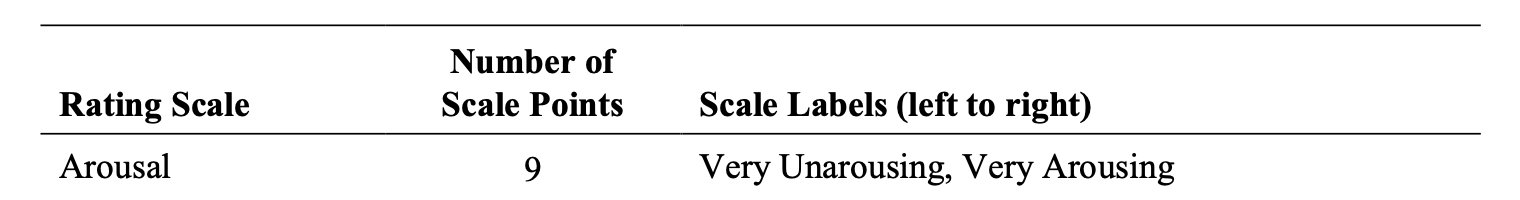} 
& Arousal is a measure of excitement versus calmness. A word is AROUSING if it makes you feel stimulated, excited, frenzied, jittery, or wide-awake. A word is UNAROUSING if it makes you feel relaxed, calm, sluggish, dull, or sleepy. 

\vskip 0.05in
Please indicate how arousing \textbf{human beings} think each word is \textbf{on a 9-point scale of VERY UNAROUSING (1) to VERY AROUSING (9), with the midpoint representing moderate arousal. Please respond using this format: word - rating}\\ \hline
Lancaster (Perceptual) \footnote{\label{Lancaster copyright}Figures reprinted from \cite{lynott2020lancaster}. Copyright 2020 by Lynott, D., Connell, L., Brysbaert, M., Brand, J., \& Carney, J. Reproduced with permission via the Creative Commons Attribution 4.0 International License (http://creativecommons.org/licenses/by/4.0/).} & You will be asked to rate how much \textbf{you} experience everyday concepts using six different perceptual senses. There are no right or wrong answers so please use your own judgement.
\vskip 0.05in
The rating scale runs from 0 (not experienced at all with that sense) to 5 (experienced greatly with that sense). \textbf{Click on a number to select a rating for each scale, then click on the Next button to move on the next item}.
\vskip 0.1in
\textbf{If you do not know the meaning of a word, just check the “Don’t know the meaning of this word" box and click "Next" to move onto the next item}. \includegraphics[width=\linewidth]{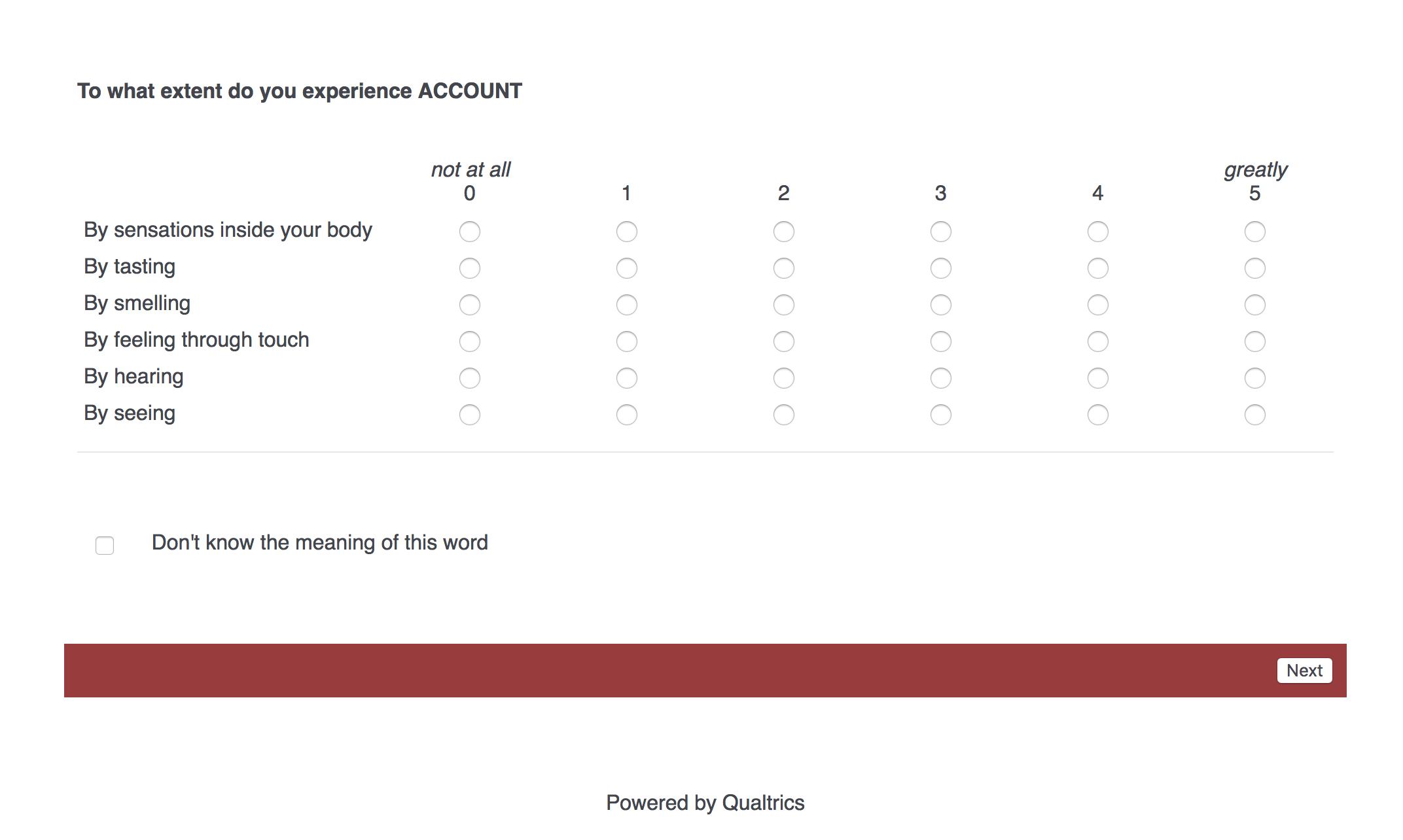} 
& You will be asked to rate how much \textbf{human beings} experience everyday concepts using six different perceptual senses. There are no right or wrong answers so please use your own judgement.
\vskip 0.05in
The rating scale runs from 0 (not experienced at all with that sense) to 5 (experienced greatly with that sense).
\vskip 0.05in
To what extent do \textbf{human beings} experience \textbf{each of the following words}:
\vskip 0.05in
By Sensations inside your body
\vskip 0.05in
By tasting
\vskip 0.05in
By smelling
\vskip 0.05in
By feeling through touch By hearing
\vskip 0.05in
By seeing

\vskip 0.05in
\textbf{Please respond using this format: word: rating, rating, rating, rating, rating (only the rating itself) and in the following order:
\vskip 0.05in
Word:rating by tasting, rating by seeing, rating by hearing, rating by smelling, rating by sensations inside your body, rating by feeling through touch}\\ \hline
Lancaster (Motor) \footnotemark[\getrefnumber{Lancaster copyright}] & You will be asked to rate how much you experience everyday concepts using actions from five different parts of the body. There are no right or wrong answers so please use your own judgment. The parts of the body will be displayed as follows:
\includegraphics[width=0.8\linewidth]{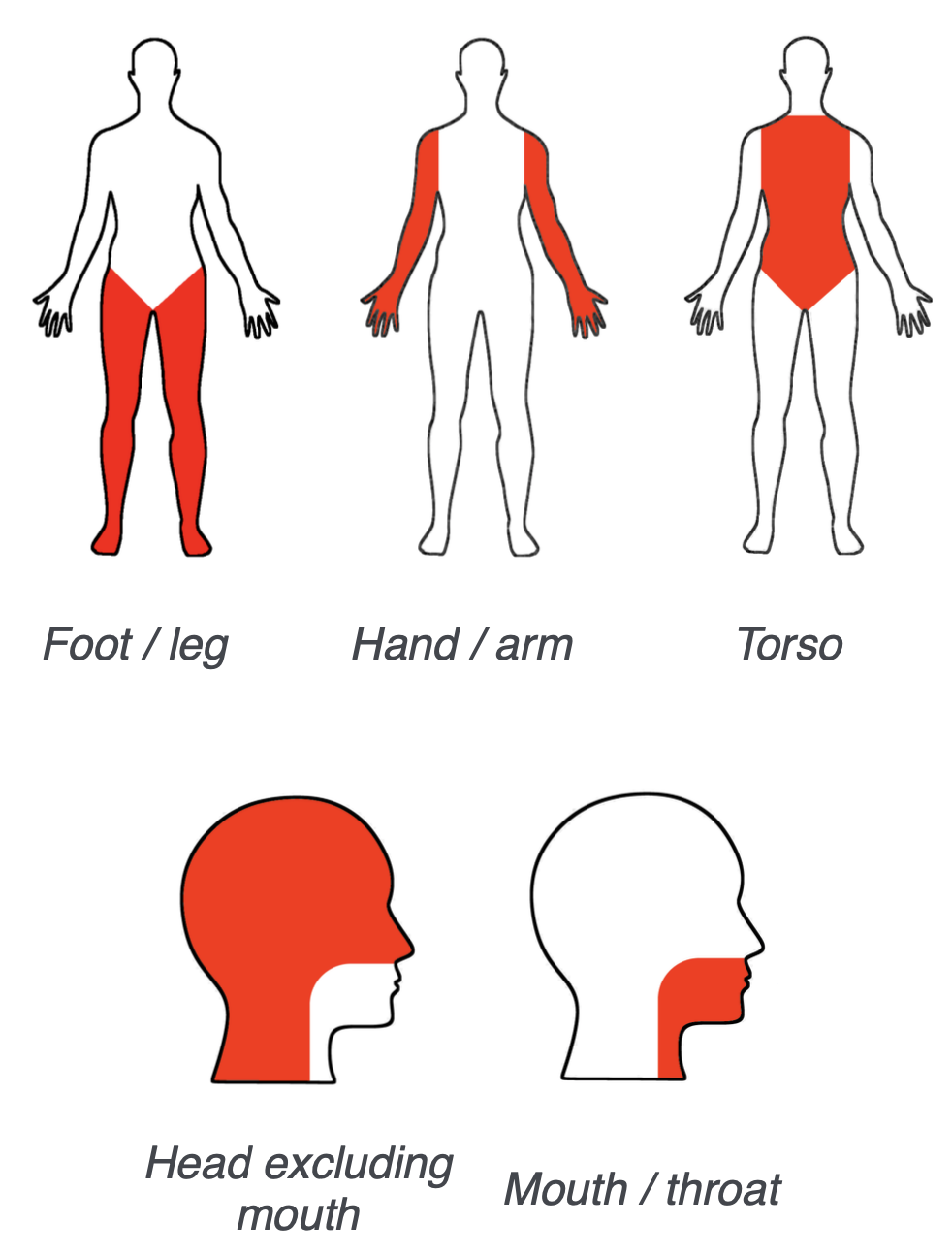}
The rating scale runs from 0 (not experienced at all with that action) to 5 (experienced greatly with that action). \textbf{Click on a number to select a rating for each scale, then click on the "Next" button to move on the next item.
\vskip 0.05in
If you do not know the meaning of a word, just check the “Don’t know the meaning of this word" box and click "Next" to move onto the next item}.
\includegraphics[width=0.8\linewidth]{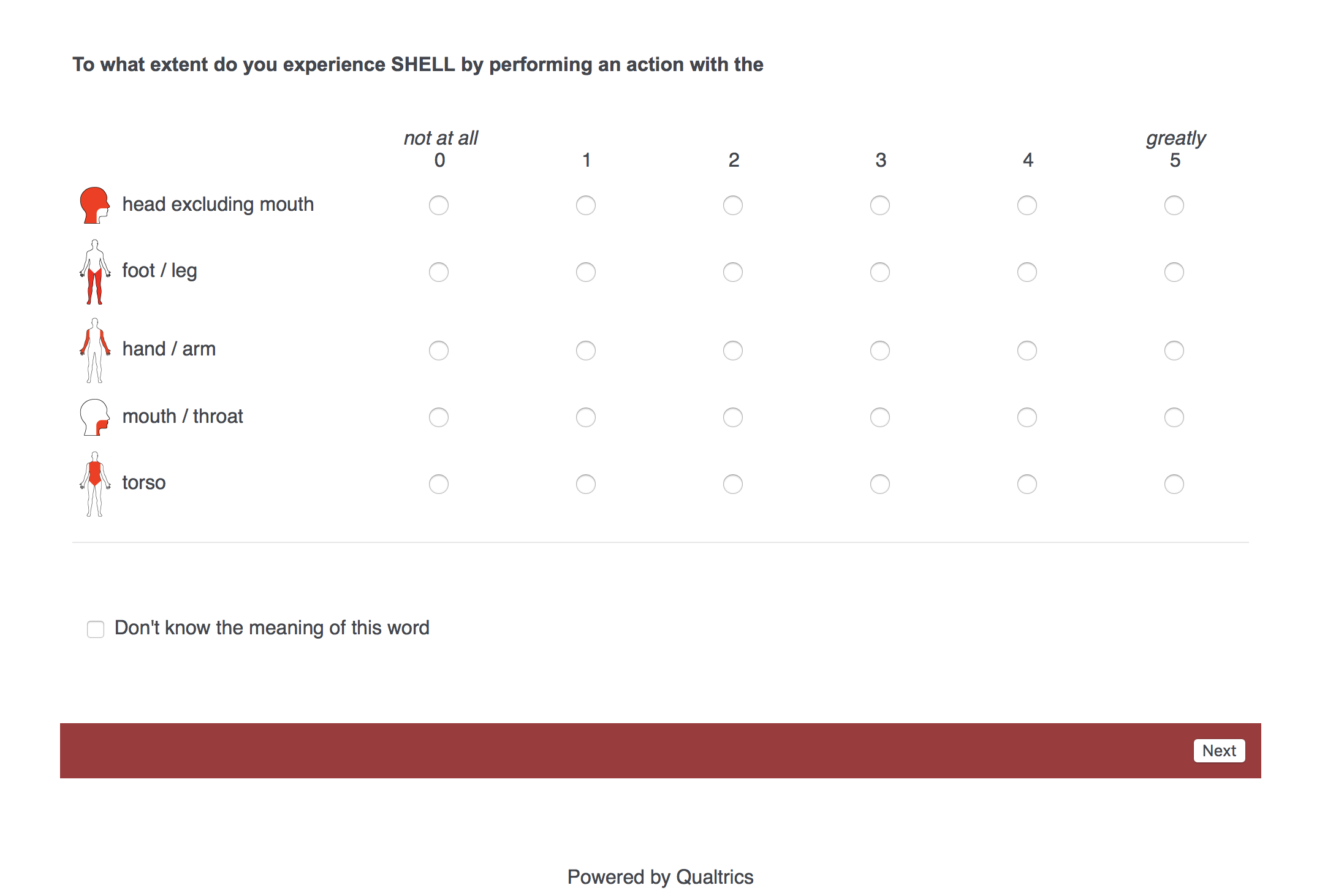} & 
You will be asked to rate how much \textbf{human beings} experience everyday concepts using actions from five different parts of the body. There are no right or wrong answers so please use your own judgment. The parts of the body will be displayed as follows:
\vskip 0.05in
\textbf{Foot / leg
\vskip 0.05in
Hand / arm
\vskip 0.05in
Torso
\vskip 0.05in
Head excluding mouth
\vskip 0.05in
Mouth / throat}
\vskip 0.05in
The rating scale runs from 0 (not experienced at all with that action) to 5 (experienced greatly with that action).
\vskip 0.05in
\textbf{Please respond using this format: word: rating, rating, rating, rating, rating (only the rating itself) and in the following order:
\vskip 0.05in
Word: Hand / arm, Foot / leg, Mouth / throat, Torso, Head excluding mouth}\\ 
\label{tab:table_s2}
\end{longtable}

\bibliographystyle{unsrt}  
\bibliography{references}